\documentclass[runningheads]{llncs}

\usepackage{eccv}
\usepackage{eccvabbrv}

\usepackage{graphicx}
\usepackage{booktabs}

\usepackage[accsupp]{axessibility}  %

\usepackage{hyperref}

\usepackage{orcidlink}

\usepackage[dvipsnames]{xcolor}
\usepackage{graphbox}
\usepackage{array}
\usepackage{pbox}
\usepackage{ifthen}
\usepackage{listofitems}
\usepackage{newfile}
\newcommand{\nsmallpts}{3889}
\newcommand{\nsmallfaces}{7774}

\DeclareMathOperator*{\argmin}{arg\,min}

\newcommand{\bu}{\ensuremath{\mathbf{u}}}
\newcommand{\bx}{\ensuremath{\mathbf{x}}}
\newcommand{\br}{\ensuremath{\mathbf{r}}}
\newcommand{\webpagelink}[1]{\href{https://remysabathier.github.io/animalavatar.github.io/}{#1} [https://remysabathier.github.io/animalavatar.github.io]}

\makeatletter
\renewcommand{\paragraph}{%
\@startsection{paragraph}{4}%
{\z@}{-0.4em}{-0.5em}%
{\normalfont\normalsize\bfseries}%
}
\makeatother

\begin{document}

\title{Animal Avatars: Reconstructing Animatable 3D Animals from Casual Videos}

\author{Remy Sabathier\inst{1,2} \and
Niloy J. Mitra\inst{2} \and
David Novotny\inst{1}}

\authorrunning{R.~Sabathier, N.~Mitra, D.~Novotny}

\institute{Meta \and
University College London}

\maketitle
\begin{abstract}
We present a method to build animatable dog avatars from monocular videos.
This is challenging as animals display a range of (unpredictable) non-rigid movements and have a variety of appearance details (e.g., fur, spots, tails).
We develop an approach that links the video frames via a 4D solution that jointly solves for animal's pose variation, and its appearance (in a canonical pose).
To this end, we significantly improve the quality of template-based shape fitting by endowing the SMAL parametric model with Continuous Surface Embeddings~(CSE), which brings image-to-mesh reprojection constaints that are denser, and thus stronger, than the previously used sparse semantic keypoint correspondences.
To model appearance, we propose an implicit duplex-mesh texture that is defined in the canonical pose, but can be deformed using SMAL pose coefficients and later rendered to enforce a photometric compatibility with the input video frames.
On the challenging CoP3D and APTv2 datasets, we demonstrate superior results (both in terms of pose estimates and predicted appearance) to existing template-free (RAC) and template-based approaches (BARC, BITE).
Video results and additional information accessible on the {\color{magenta}\webpagelink{\texttt{project page}}}.

\end{abstract}

\section{Introduction}
\label{sec:intro}
Building poseable reconstructions of animals captured by consumer imaging devices is a valuable technology with numerous applications in augmented and virtual reality.
Among many possible animal species, the reconstruction of canines is of particular interest primarily due to their important role in the lives of their two-legged friends.

While it is nowadays possible to reliably reconstruct rigid scenes captured from a moving vantage point \cite{schonberger2016structure}, the reconstruction of non-rigid shapes is significantly less constrained and, as such, a more challenging problem.
Here, many recent works focused on generic animal reconstruction from multi-view (videos) \cite{Kokkinos_2021_CVPR,yang_lasr_2021,yang_banmo_2021} or single-view 2D image supervision  \cite{wu2023magicpony} without prior knowledge of the animal shape.
While the latter demonstrated impressive progress, the challenging nature of the problem limits the applicability to scenarios with simple deformations and good multi-view test-time coverage.

\begin{figure}[t!]
   \centering
\captionsetup{type=figure}
\includegraphics[width=.97\textwidth]{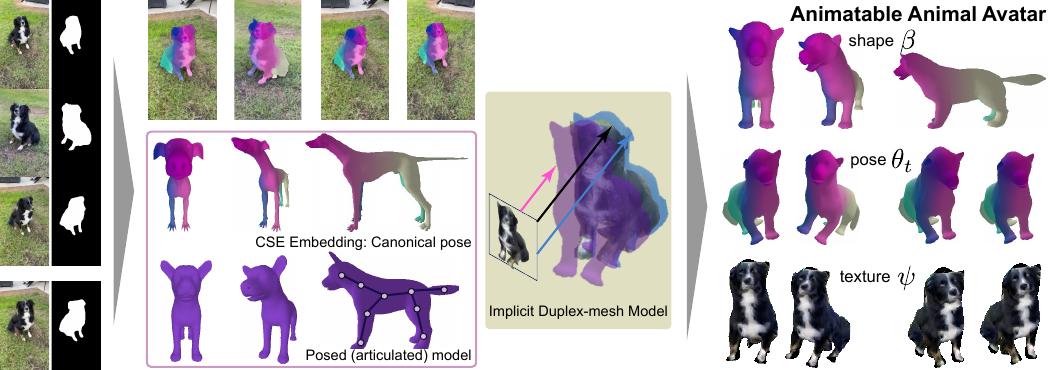}
\captionof{figure}{
\textbf{Animal Avatars}. Given a monocular video of a dog, we propose a template-based method to reconstruct the shape $\beta$, motion $\theta_{t}$ and texture $\psi$. We address the challenge of insufficient supervision for unconventional views by integrating Continuous Surface Embeddings with an articulated mesh. We introduce a novel implicit duplex-mesh texture model, jointly optimized alongside motion parameters.}
\label{fig:teaser-fig}
\end{figure}

To better constrain the non-rigid 3D reconstruction, inspired by state-of-the-art methods for Human 3D reconstruction~\cite{loper_smpl_2015,dong2023ag3d}, we leverage a known category template. 
Specifically, we use the SMAL template introduced by 
Zuffi \etal~\cite{zuffi_3d_2017} - a quadruped-equivalent of the seminal SMPL parametric human model \cite{loper_smpl_2015}.
The template has been used to enable single-view 3D dog reconstruction trained on an extensive collection of dog images semi-manually annotated with SMAL poses \cite{rueegg_barc_2022,ruegg_bite_2023,ehsani_who_2018}.
While this approach provides a clear state-of-the-art in monocular 3D dog-shape reconstruction, the inherent ambiguity of single-view reconstruction provides many challenges.

To further increase our chances of successful reconstruction, besides leveraging the SMAL model, we turn our attention towards reconstructing casual video captures of dogs, which was first explored in \cite{biggs_creatures_2018}.
Indeed, videos provide a stronger multi-view supervision which significantly simplifies the 3D shape fitting problem.
However, regardless of monocular \cite{rueegg_barc_2022,ruegg_bite_2023} or multi-view conditioning \cite{biggs_creatures_2018}, we observed a common failure mode in existing methods when animals are viewed from non-frontal views.
This issue arises because fitting relies on sparse joint-reprojection constraints that mainly cover front-facing parts, offering limited supervision for rear and side views.

Hence, our first contribution is to replace the sparse keypoint supervision with a denser alternative.
Specifically, we exploit Continuous Surface Embeddings \cite{neverova_continuous_2020} (CSE), which annotate \emph{each} vertex of the CSE dog mesh with a unique descriptor.
Furthermore, CSE provides a pre-trained deep image-to-CSE predictor, that labels image pixels with their corresponding CSE descriptor and transitively with the matching mesh vertex.
In this work, in an one-time process, we first transform CSE descriptors to the SMAL mesh by means of a semi-manual non-rigid aligment.
This then enables a stronger keypoint loss providing reprojection constraints for all points on the animal's body, even in rear views.

Secondly, we enhance fits by exploiting the inherent smoothness of animal movements over time.
Previous attempts incorporated this knowledge by enforcing temporal smoothness on the deformation coefficients of the SMAL template.
However, we found this approach flawed because the coefficients have to represent both the smooth animal motion and the camera motion, which is often non-smooth due to the unstable camera operator. 
Instead, we propose to represent the SMAL deformation as a combination of accurate Structure-from-Motion camera and optimized animal motion, allowing for proper temporal regularization.
Importantly, SfM also provides intrinsic parameters of each camera (focal length), which facilitates more accurate shape fitting through loss terms that require rendering.

Finally, we are the first to enable texturing of the SMAL mesh by leveraging it as a scaffold for an implicit duplex-mesh neural radiance field, which can be rendered while accounting for body deformations. 
Our approach involves defining implicit shape and color functions on a subset of the 3D domain bounded by enlarged and downsized versions of the mesh template.
This allows for articulation of the corresponding implicit surface by posing the boundary meshes similarly to the original mesh.

We evaluate the color and 3D shape-fitting accuracy on the CoP3D dataset \cite{sinha_common_2022}, containing crowd-sourced ``turntable'' videos of dogs captured by smartphone cameras, achieving performance superior to template-based \cite{rueegg_barc_2022,ruegg_bite_2023} and template-free \cite{yang_reconstructing_2023} baselines. We also compare our pose estimation quality on the recent APTv2 dataset~\cite{yang2023aptv2}
and report results superior compared to video-based methods dog-specific~\cite{yang_reconstructing_2023}. 

\section{Related Work}

\paragraph{Video reconstruction on humans.}

    Recent works in 3D human pose reconstruction show impressive results on videos, representing detailed motions and robustness to occlusions \cite{goel_humans_2023, zhang_pymaf-x_2023}. Several factors support the progress in this area.

    The majority of methods rely on the parametric \emph{SMPL} model introduced in \cite{loper_smpl_2015} and refined in \cite{romero_embodied_2017, pavlakos_coarse--fine_2017, osman_star_2020}. Unlike the existing animal models, it is learned from a large collection of real 3D scans of humans, which entails stronger expressiveness. The model provides parametric handles to both shape and pose variations. 
    Additionally, human-centric models benefit from large real datasets with keypoint annotations \cite{lin_microsoft_2015,wu_ai_2019, mehta_monocular_2017} as well as synthetic 3D datasets \cite{ionescu_human36m_2014}.

    We also note the availability of off-the-shelf models for related tasks such as human key-point identification \cite{xu_vitpose_2023}.
    Such models can guide training of 3D reconstruction or provide soft annotation on large unlabelled datasets \cite{gu_ava_2018}.
    These factors explain why 3D animal reconstruction cannot directly benefit from breakthroughs in the human domain.

    Additionally, there are several works \cite{chen2021animatable, peng2021animatable, jiang2022neuman, guo2023vid2avatar} targeting human reconstruction with texture from monocular and multi-view sources achieving high rendering quality by leveraging off-the-shelf human pose estimation models.

\paragraph{Template-based animal reconstruction.}
Based on the success of template-based human reconstruction using SMPL, \cite{zuffi_3d_2017} introduced \textit{SMAL}, a parametric model for quadruped animals. 
Unlike the SMPL model, which is supervised by scans of real humans, and due to the many challenges of scanning live quadruped animals, the SMAL model is only trained with scans of toy animal models.

To add to the challenge, no large-scale dataset with 3D annotations for dog reconstructions exists. 
The only 3D-annotated dataset \cite{kearney_rgbd-dog_2020} does not adequately represent the diversity of dog breeds and poses. 
The most relevant image dataset is \emph{StanfordExtra}~\cite{ehsani_who_2018,khosla_novel_2011}, a collection of dog images with silhouette and joint annotations. 
Despite being diverse, spanning different dog breeds and environments, the dataset is biased towards front-facing views.
This motivates our choice of the CoP3D dataset \cite{sinha_common_2022}, an extensive collection of pet videos with high view-point variability in each video.

Similar to human reconstruction research, multiple works estimate shape attributes $\beta$ and pose attributes $\theta$ for the SMAL model from a \textit{single} image, relying on 2D reprojection constraints. 
\cite{biggs_creatures_2018} predicts skeleton joints to find an initial solution, which is then refined to match keypoints estimated from the animal silhouette. \cite{li2021coarsetofine} propose a coarse-to-fine strategy, where an initial solution is refined through a graph-convolutional network. BARC~\cite{rueegg_barc_2022} enforces similar shape attributes for dogs of the same \emph{breed}, which is predicted by a deep network. BITE~\cite{ruegg_bite_2023}, an extension of BARC, improve prediction plausibility with \emph{ground-contact} and \emph{ground-plane} losses, and improve accuracy with an iterative refinement loop. We compare ours against BARC and BITE to show how multi-view supervision and CSE embeddings significantly improve performance, especially on the challenging videos from the COP3D dataset. We note some additional related works on other species \cite{badger2020, wang2021birds}.

\paragraph{Template-free animal reconstruction.}
The template-free setup enabling reconstruction of a wider range of animals has also been considered.
These approaches build a 3D representation by analyzing a collection of images, or frames of a single video, or videos of the same species.

Progress made in differentiable rendering \cite{ravi_accelerating_2020,liu_soft_2019,wiles2020synsin} supported the \emph{analysis-by-synthesis} method for animal reconstruction from a single video. Recent works  
\cite{yang_lasr_2021,Kokkinos_2021_CVPR,novotny2022keytr,yang2021viser} propose to minimize silhouette and photometric losses in order to jointly learn camera parameters and a textured frame-independent linear deformation 3D model. 
We are inspired by Viser~\cite{yang2021viser} that recovered articulated humans from monocular videos. 
They learn long-range 2D point tracks using object masks and optical flow, and a video-specific embedding linking pixel appearance to surface points on a canonical deformable mesh. 
However, such approaches are vulnerable to (self-)occlusions, especially in significantly dynamic scenes.

To overcome these issues, several works train on a set of videos. \emph{BANMo} \cite{yang_banmo_2021} leverages multiple videos of the same subject to build an implicit canonical representation, posed via a differentiable \emph{neural blend skinning} method. Similar to our work, they use pixel \emph{CSE} predictions on the different images to link it with a 3D embedding defined in the canonical space. 
As such, satisfactory new views are generated only when positioned relatively close to the training views.
Recently, RAC~\cite{yang_reconstructing_2023} extended BANMo to learn a general and instance-specific model from a set of videos from the same category of deformable objects, including the dog category analyzed here.
TrackeRF \cite{sinha_common_2022} extends PixelNeRF \cite{yu_pixelnerf_2020} to time-deforming shapes but it only predicts novel views without a full-scale 3D animal-body model.
Note that the method additionally needs to be initialized with a basic hierarchical joint skeleton.
We compare with RAC in \cref{sec:experiments} and omit comparison to TrackeRF because its source code is not available.

\begin{figure*}[t!]
    \centering
    \includegraphics[width=\textwidth]{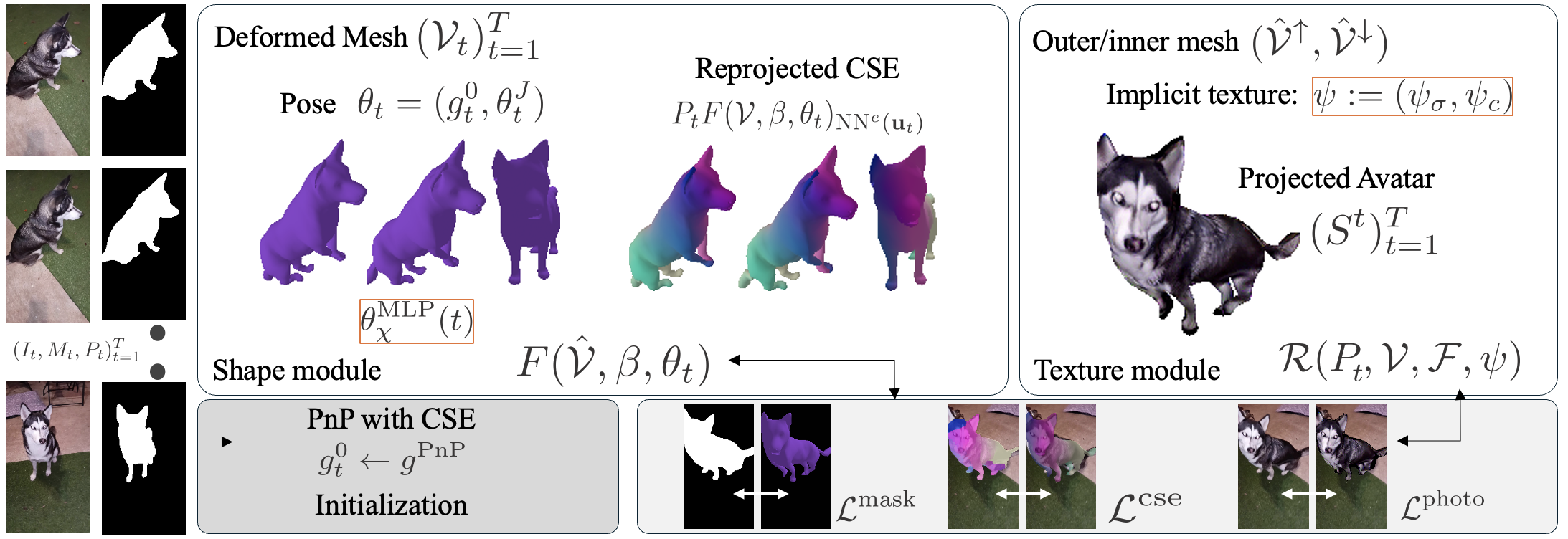}
    \caption
    {\textbf{Method overview}. Two stage process: \textit{First}, we initialize root pose $g_{t}^{0}$ via PnP-RANSAC, utilizing CSE mesh-pixel correspondences. \textit{Then}, we jointly optimize shape $\beta$, time-varying pose $\theta_{t}$ and implicit texture $\psi$ through an analysis-by-synthesis approach, leveraging mask $\mathcal{L}^{mask}$, dense correspondence $\mathcal{L}^{cse}$ (optimized models in orange), and     
    photometric $\mathcal{L}^{photo}$ signals. 
    }
    \label{fig:method_overview}
\end{figure*}

\section{Method} \label{sec:method}
Our goal is to reconstruct the 3D shape and appearance of a dog captured in a monocular video, i.e., given a tuple $(I_t)_{t=1}^{T}$ of $T \in \mathbb{N}$ video frames
we output a tuple $(S_t)_{t=1}^{T}$ of colored 3D shapes $S_t$ of the animal in each frame $I_t \in \mathbb{R}^{3 \times H \times W}$.

In \cref{sec:representation}, we detail the shape representation $S$,
while \cref{sec:pose_init,sec:shape_fitting} describe the optimization process recovering the shape $S$ given input images $I$.

\subsection{Shape and Appearance Representation} \label{sec:representation}
Our method defines the colored shape representation $S_t$ at time $t$ as a 3-tuple
\begin{equation} \label{eq:main_shape_rep}
S_t := (\beta, \theta_t, \psi),
\end{equation}
where $\beta$ and $\psi$ define the intrinsic (i.e., time-invariant) deformation and the texture of the dog body respectively,
and $\theta_t$ is the time-dependent pose of its skeleton.
In what follows, we detail these three sets of parameters.

\paragraph{3D shape representation $\beta,\theta_t$.}
There is a plethora of articulated 3D shape representations ranging from universal less-constrained 3D-flow functions \cite{pumarola2021d,gao2021dynamic} to category-specific Linear-Blend-Skinning (LBS) models attached to a fixed surface-mesh template \cite{lewis2023pose,loper_smpl_2015,zuffi_3d_2017}.
Since we focus on a certain animal category (i.e., dog) with a well-defined space of plausible articulations and body deformations, we opt for the latter, i.e., a template-based shape model. 

Specifically, we represent the 3D geometry (i.e., the parameters $\theta_t, \beta$ in \cref{eq:main_shape_rep}) of each dog with the SMAL model~\cite{zuffi_3d_2017}.
The latter comprises a deformable template mesh $(\hat{\mathcal{V}}, \mathcal{F})$ with vertices $\hat{\mathcal{V}} \in \mathbb{R}^{\nsmallpts \times 3}$ and a list of triangular faces $\mathcal{F} \in \mathbb{N}^{\nsmallfaces \times 3}$.
The deformation of $\hat{\mathcal{V}}$ is defined as a function 
\begin{equation} \label{eq:smal_deformation}
F(\hat{\mathcal{V}}, \beta, \theta_t) := \mathcal{V} \in \mathbb{R}^{\nsmallpts \times 3},
\end{equation}
conditioned on shape parameters (PCA coefficients and bone lengths) $\beta \in \mathbb{R}^{d_\beta}$, controlling the non-rigid deformation of the animal shape in its \textit{canonical} pose, and pose parameters $\theta_t := (g^0, \theta^J)$.
The latter has two components: (i)~a root transformation $g^0 \in \mathbb{SE}(3)$ that represents the overall rigid transformation of the dog body; and (ii)~angles $\theta^J \in \mathbb{R}^{d_J}$ of the animal joints that control the deformation of its limbs.

\paragraph{Time-deforming SMAL.}
To represent the time-varying deformation of a dog, we estimate a tuple $(\theta_t)_{t=1}^{T}$ comprising SMAL pose coefficients $\theta_t$ for each of the $T$ video frames, and a single vector 
$\beta$ because, typically, the intrinsic deformation is time-invariant.
Since the animals often move smoothly in time, the pose $\theta_t$ is defined as a function 
$\theta^\text{MLP}(t)$
of a smooth temporal basis $\gamma(\tau(t))$ as follows:
\begin{equation}
\theta_t := \theta^\text{MLP}(\tau(t)),
\end{equation}
implemented by a shallow multi-layer perceptron $\theta^\text{MLP}$ accepting
positional encoding $\gamma$ \cite{vaswani_attention_2017} applied to the timestamp $\tau(t) \in \mathbb{R}^+$ of frame $I_t$.

\begin{figure}[t]
    \centering
    \newcommand{\csekpfigsize}{1.5cm}
\graphicspath{{figures/img_folder/cse/}{figures/img_folder/}}

\newcommand*{\groupfig}[2]{%
\includegraphics[height=\csekpfigsize,width=\csekpfigsize,keepaspectratio,align=c]{#1_#2_frame_0}&%
\includegraphics[height=\csekpfigsize,width=\csekpfigsize,keepaspectratio,align=c]{#1_#2_frame_150}&%
\includegraphics[height=\csekpfigsize,width=\csekpfigsize,keepaspectratio,align=c]{#1_#2_frame_160}%
}

\newcolumntype{C}{>{\centering\arraybackslash}m{1cm}}

\newcommand*{\rowlabelcse}{%
\footnotesize%
\phantom{a}%
\rotatebox{90}{\hspace{-0.1cm}CSE}%
}

\newcommand*{\rowlabelkp}{%
\footnotesize%
\phantom{a}%
\rotatebox{90}{\hspace{-0.6cm}Keypoints}%
}

\renewcommand{\arraystretch}{0}
\setlength{\tabcolsep}{0pt}
\centering%

\begin{tabular}{cc}

\begin{tabular}{cc}%
\footnotesize CSE template \cite{neverova_continuous_2020} \phantom{a} \\
\includegraphics[height=2.4cm,width=3cm,keepaspectratio,align=c]{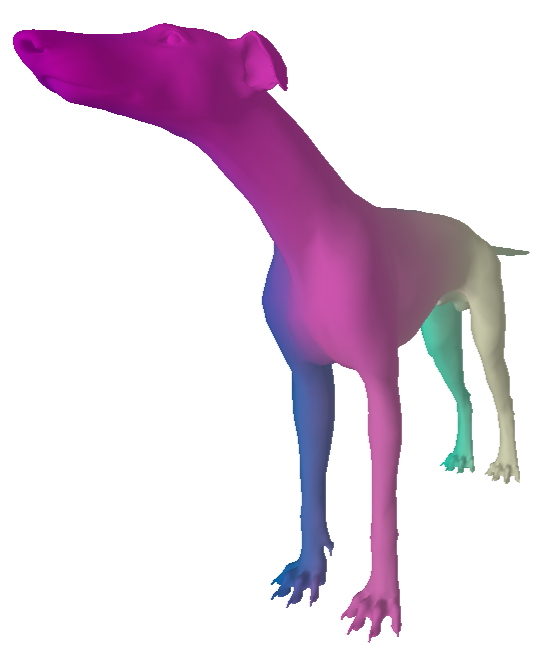} \\%
\parbox{2.9 cm}{%
    \centering%
    \footnotesize%
    \vspace{0.2cm}%
    $\downarrow$ \textit{Mesh alignment} $\downarrow$%
    \vspace{0.2cm}\\
} \\%
\footnotesize SMAL template \cite{zuffi_3d_2017} \phantom{a}\\
\includegraphics[height=2.4cm,width=3cm,keepaspectratio,align=c]{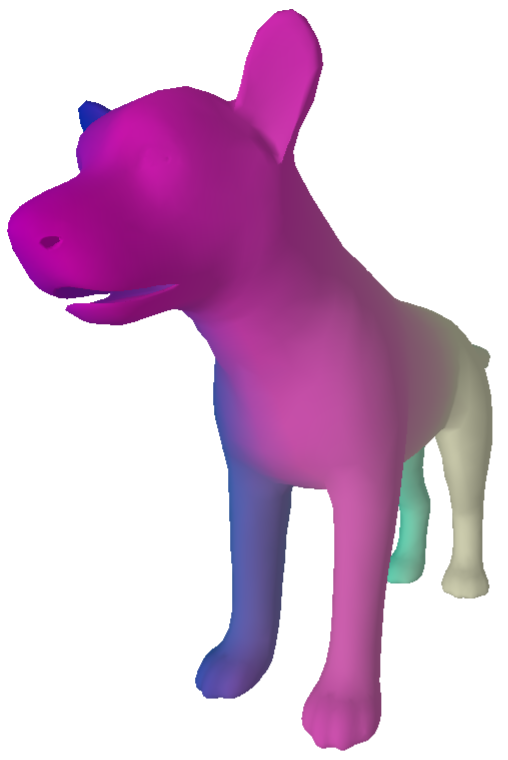}%
\end{tabular} &%

\begin{tabular}{cccccccccccc}%
\groupfig{cse}{600_92491_183070}&\groupfig{cse}{1028_21488_15551}&\groupfig{cse}{1054_47075_43215}&\rowlabelcse\\
\groupfig{bite_kp}{600_92491_183070}&\groupfig{bite_kp}{1028_21488_15551}&\groupfig{bite_kp}{1054_47075_43215}&\rowlabelkp\\
\groupfig{cse}{1033_26269_19535}&\groupfig{cse}{1043_36452_31495}&\groupfig{cse}{1043_36548_32210}&\rowlabelcse\\
\groupfig{bite_kp}{1033_26269_19535}&\groupfig{bite_kp}{1043_36452_31495}&\groupfig{bite_kp}{1043_36548_32210}&\rowlabelkp\\
\end{tabular}

\end{tabular}

\let\groupfig\relax
\let\rowlabelkp\relax
\let\rowlabelcse\relax
\let\csekpfigsize\relax

    \caption{\textbf{SMAL CSE}. We align the CSE mesh template (top left) with the SMAL template (top right) in order to setup the CSE coordinate frame over the surface of the SMAL mesh.
    In combination with a pretrained image-to-CSE predictor, this allows establishing dense correspondences between surface points of the SMAL template and the corresponding image pixels.
    Note that the image CSE detections (rows labelled ``CSE'') provide dense correspondence covering all parts of the dog body, which is not the case of sparse keypoints (``Keypoints'' rows).
    }
    \vspace{0.2in}
    \label{fig:smal_cse}
\end{figure}

\paragraph{Continuous SMAL Embeddings.}
Besides the deformation parameters ($\theta, \beta$), SMAL also defines joint locations $\hat{\mathcal{J}} \in \mathbb{R}^{N_J \times 3}$ as a sparse subset of 3D surface points, 
which are linearly regressed from the vertex locations.
Since the joints correspond to semantic parts of the animal body (paws, ear tips), they can improve shape-fitting accuracy via establishing correspondences between the SMAL mesh and the detections of the body parts in the images.
However, while these keypoints improve performance in \cite{rueegg_barc_2022,ruegg_bite_2023}, where most animals are photographed from their side or frontal views, they are inadequate for our video fly-arounds containing rear views with little-to-no visible keypoints (see \cref{fig:smal_cse} and~\cref{sec:experiments}).

Hence, instead of sparse landmarks, we exploit Continuous Surface Embeddings (CSE) \cite{neverova_continuous_2020}, which attach a unique embedding vector $e_k \in \mathbb{R}^{d_e}$ to \emph{each} vertex $X_k \in \hat{\mathcal{V}}$ of the (canonical) SMAL template such that the dimensions of $e$ vary smoothly over the mesh surface.
CSE also provides a deep predictor that annotates each image pixel with its corresponding mesh coordinate $e_k$.
Thus, unlike sparse keypoints, the latter densely annotates images of animals from arbitrary viewpoints, including rear views.

Because the original template mesh of CSE \cite{neverova_continuous_2020} is different from the SMAL template, following \cite{neverova_continuous_2020}, we transform the CSE coordinate map to SMAL using a customized variant of the Zoom-Out method \cite{melzi_zoomout_2019}.
The latter results in the final set of SMAL-CSE vertex coordinates $e_k$ (alignment visualized in \cref{fig:smal_cse}).

\iftrue  %

\paragraph{Double-mesh implicit texture $\psi$.}
Besides reconstructing 3D animal shapes, we also aim to recover the texture of the animal body.
We require an exact supporting 3D shape to learn an accurate texture model.
However, due to the low expressivity of the SMAL deformation space, the posed mesh can only represent the surface of a dog up to a certain error.
Thus, following recent advances in new-view synthesis of humans \cite{lombardi_neural_2019}, we leverage the template mesh as a scaffold supporting a more accurate implicit radiance field \cite{mildenhall_nerf_2020}, which we describe next.

Our method is inspired by duplex radiance fields \cite{wan2023learning}, originally proposed for speeding up rendering of neural radiance fields \cite{mildenhall_nerf_2020}.
Specifically, we first extrude the 2-manifold canonical surface to a 3D volume by defining an $\epsilon$-neighborhood $\hat{\mathcal{N}}^\epsilon \subset \mathbb{R}^3$ as a 3D subspace bounded by an \textit{outer} mesh with vertices $\hat{\mathcal{V}}^\uparrow = (1 + \epsilon) \hat{\mathcal{V}}$, and an \textit{inner} mesh $\hat{\mathcal{V}}^\downarrow = (1 - \epsilon) \hat{\mathcal{V}}$, both sharing the same faces $\mathcal{F}$.
In practice, we obtain the offset meshes by moving along directions of vertex normals. 
Similar to the template mesh itself, both $\hat{\mathcal{V}}^\uparrow$ and $\hat{\mathcal{V}}^\downarrow$ can be deformed with $\theta, \beta$ resulting in the posed neighborhood $\mathcal{N}^\epsilon$ bounded by $F(\hat{\mathcal{V}}^\uparrow, \beta, \theta)$ and $F(\hat{\mathcal{V}}^\downarrow, \beta, \theta)$.
    
Within this neighborhood, we then define an opacity function $\psi_\sigma: \mathcal{N}^\epsilon \mapsto \mathbb{R}^+$, annotating 3D locations $\hat{\bx}$ in the mid-space with presence ($\psi_\sigma(\hat{\bx}) > 0$) or absence ($\psi_\sigma(\hat{\bx}) \approx 0$) of the surface, and the radiance function
$\psi_c: \mathcal{N}^\epsilon \times \mathbb{S}^2 \mapsto[0, 1]^3$,
which colors the space depending on the direction $\br \in \mathbb{S}^2$ from which the input point $\hat{\bx}$ is observed.
The coloring and opacity functions are implemented as in \cite{chan2022efficient}, i.e., using a shallow MLP decoding a learnable triplane feature grid.
Please refer to the supplementary material for details.

\begin{figure}[h!]
    \centering
\includegraphics[width=0.9\columnwidth]{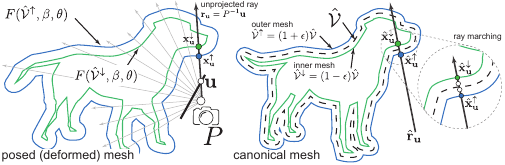}
    \caption{%
    \textbf{Implicit duplex-mesh model.}
    We propose a novel deformable implicit shape model.
    Using the radiance and opacity functions $\psi_c$ and $\psi_\sigma$ defined over an $\mathbb{R}^3$ subspace bounded by a canonical duplex mesh with vertices $\mathcal{\hat{V}}^\uparrow$ and $\mathcal{\hat{V}}^\downarrow$, we render a color of the posed duplex mesh via EA raymarching over a canononical ray $\hat{\br}_\bu$.
    The latter is obtained by transforming into the canonical space the intersections of the view-space ray $\br_\bu$ with the posed duplex mesh $F(\mathcal{\hat{V}}^\uparrow, \beta, \theta), F(\mathcal{\hat{V}}^\downarrow, \beta, \theta)$.
    }
    \label{fig:fast plane}
\end{figure}

\paragraph{Double-mesh rendering.}
Having defined the animal shape $(\mathcal{V}, \mathcal{F})$ and the implicit texture $\psi$, we image the latter from an arbitrary camera viewpoint $P$ using a differentiable rendering function $\mathcal{R}$:
\begin{equation}
    \mathcal{R}(P, \mathcal{V}, \mathcal{F}, \psi) := \bar{I},
    \end{equation}
which outputs the render $\bar{I} \in [0, 1]^{3 \times H \times W}$ as observed from the camera with projection matrix $P \subset \mathbb{R}^{3 \times 4}$.
    
To obtain $\bar{I}$, we iterate over all its pixels $\bu \in [1..H] \times [1..W]$ and march with Emission-Absorption~(EA) over the canonical ray $\hat{\br}_\bu$ defined as the camera-space ray $\br_\bu = P^{-1} \bu$ in the rest-pose coordinates.
Specifically, $\br_\bu$ is first intersected with the posed outer boundary mesh $F(\hat{\mathcal{V}}^\uparrow, \beta, \theta)$, and then the intersection's barycentric coordinates are applied to the corresponding canonical mesh $\hat{\mathcal{V}}^\uparrow$ yielding a 3D point $\hat{\bx}_\bu^\uparrow \in \hat{\mathcal{N}}^\epsilon$ in the canonical neighborhood $\hat{\mathcal{N}}^\epsilon$.
Repeating the same for the for the inner mesh $F(\hat{\mathcal{V}}^\uparrow, \beta, \theta)$ yields a second 3D point $\hat{\bx}^\downarrow_\bu \in \hat{\mathcal{N}}^\epsilon$.
The two points $\hat{\bx}_\bu^\uparrow$ and $\hat{\bx}^\downarrow_\bu$ then define the canonical ray $\hat{\br}_\bu$, over which we march with EA, accumulating the outputs of the radiance functions $\psi_c$ and $\psi_\sigma$ in the process, to render the final color of pixel $\bu$ (details in the supplementary).

Note that our EA rendering differs from the duplex radiance fields \cite{wan2023learning}, which instead leverage an MLP to directly map positional encodings of the two intersection points  $\hat{\bx}^\downarrow_\bu$, $\hat{\bx}_\bu^\uparrow$ to a surface color.

\else  %
    
\paragraph{Appearance representation $\psi$}
Besides reconstructing 3D animal shape, our work also aims to recover the texture of the animal body.
To do so, we describe the mesh color via a time-invariant coloring function
$c: \mathbb{R}^{d_e} \mapsto[0, 1]^3$ 
    assigning to each coordinate $e \in \mathbb{R}^{d_e}$ on the surface of the mesh $(\mathcal{V}, \mathcal{F})$ a three-dimensional RGB color $c(e) \in [0, 1]^3$.
    Here, $c$ is implemented as a \todo{two}-layer perceptron
    \begin{equation}
    c^\text{MLP}_\psi(e) = \text{MLP}_\psi(\gamma(e)),
    \end{equation}
    with parameters $\psi$ accepting the positional encoding $\gamma(e)$ \cite{mildenhall_nerf_2020} of CSE coordinates $e$.
    Note that this is convenient because we are effectively representing the surface-smooth texture as a function of the similarly-smooth coordinates $e$.
    
    \paragraph{Shape rendering}
    Having defined the animal shape $(\mathcal{V}, \mathcal{F})$ and texture $c$, we image the latter from an arbitrary camera viewpoint $P$ using a differentiable rendering function $R$:
    \begin{equation}
    R(P, \mathcal{V}, \mathcal{F}, \psi) = \bar{I},
    \end{equation}
    which outputs the render $\bar{I} \in [0, 1]^{3 \times H \times W}$ as observed from the camera with projection matrix $P \subset \mathbb{R}^{3 \times 4}$.
    Along with the latter, $R$ takes as input the SMAL faces $\mathcal{F}$ and the deformed vertices $\mathcal{V} = (\hat{\mathcal{V}}, \beta, \theta)$ of the SMAL template.
    
    $R$ also accepts parameters $\psi$ of the appearance MLP $c^\text{MLP}_\psi$ which define the rendered colors.
    To render the color image $\bar{I}$, we first output a CSE render via barycentric interpolation of the SMAL-CSE vertex texture $\{e_k\}_{k=1}^{\nsmallpts}$ using the rendered fragment buffer, followed by evaluating the coloring MLP $c^\text{MLP}_\psi$ at each pixel of the CSE render.
    We implement $R$ using the differentiable mesh rasterizer from PyTorch3D \cite{ravi_accelerating_2020}.

\fi

\subsection{Pose Initialization} \label{sec:pose_init}
Fitting a non-rigid shape to a monocular video is a challenging task and, as such, a trivial random initialization of the shape parameters (the weights of MLPs $\psi, \chi$) inevitably leads to a failure.
We thus employ two fitting stages, where the first initializes parameters to ensure convergence of the second stage.

\paragraph{Root pose estimation.}
We observed that a suitable initialization of the root pose $g^0$, while initializing the rest of the parameters (limb angles $\theta^J$, intrinsic deformation $\beta$, implicit MLP $\psi$) randomly, is sufficient to avoid the most common local minima, such as flipping of the dog body along its symmetry axes.

The goal of the initial fitting stage is thus to recover the root rigid transformations $g^0_t$ for $t\in[1,T]$ so that the perspective projection of the unposed canonical template $\hat{\mathcal{V}}$ into each camera $P_t$ matches the depicted dog in frame $t$.

\paragraph{PnP with CSE.}
To this end, we leverage the CSE coordinate map of the SMAL mesh (\cref{sec:representation}).
First, for each image $I_t$, a pre-trained CSE convolutional network annotates pixels $\mathbf{u}_t \in [1..H]\times[1..W]$ with their CSE embedding $e(\mathbf{u}_t)$.
Then, we establish correspondences between each pixel $\mathbf{u}_t$ and the vertices $\hat{X}_{\text{NN}^e(\mathbf{u}_t)} \in \hat{\mathcal{V}}$ on the template mesh $\hat{\mathcal{V}}$ by recovering the nearest neighbors
$\text{NN}^e(\bu_t) := \argmin_{1 \leq k \leq |\mathcal{V}|} \| e(\mathbf{u}_t) - e_k \|$
in the CSE embedding space.
Given the set
$\{(\bu_t, \hat{X}_{\text{NN}^e(\bu_t)})\}$
of pixel-to-vertex correspondences, PnP-RANSAC \cite{efficientPnp_09} estimates the best camera $P_t^\text{PnP}$ aligning the projections of the vertex points with their corresponding pixels.

In order to increase robustness to occasional failures of PnP caused by inaccurate CSE predictions, we employ a collective pose refinement that finds a single global rigid transformation $g^\text{PnP} \in \mathbb{SE}(3)$ aligning the sequence of PnP-estimated cameras $(P_t^\text{PnP})_{t=1}^T$ with the scene SfM cameras $(P_t^\text{SfM})_{t=1}^T$ (the SfM cameras are detailed in the next section).
The latter then initializes the root-rigid transformation of each frame, i.e., $\forall t: g^0_t = g^\text{PnP}$.

\subsection{Shape Fitting} \label{sec:shape_fitting}
We now detail the second fine-fitting stage, which optimizes all shape parameters ($\beta, \theta_t, \psi$) given the initial poses $g^0$.

\paragraph{Factoring the rigid pose.}
Even with near-perfect initialization of the root rigid pose, it is challenging to converge to a good solution when, as done in previous works \cite{ruegg_bite_2023,rueegg_barc_2022}, the rigid component $g_t = g^0_t \in \mathbb{SE}(3)$ of the rendering camera $P_t$ is solely represented with the root transformation $g^0_t$.
Such optimization is challenging because $g^0_t$ has to represent the animal pose and also compensate for complex camera motions such as the jitter caused by unstable handling.

Instead, we factor the extrinsics $g_t = g^\text{cam}_t \cdot g^0_t$ of each rendering camera $P_t$ as a composition of the camera motion $g^\text{cam}_t \in \mathbb{SE}(3)$ w.r.t. the rigid scene background and the motion $g^0_t \in \mathbb{SE}(3)$ of the dog w.r.t. the background.
Here, $g^\text{cam}_t := g^\text{SfM}_t$ is fixed to the SfM camera $g^\text{SfM}_t$ estimated by COLMAP~\cite{schonberger2016structure} which we empirically found to be very accurate.
Offloading the camera estimate to SfM, we are then left with the simpler task of regressing the temporally-smooth rigid animal motion $g^0_t$.

\paragraph{CSE-guided fine shape fitting.}
The second fine fitting stage outputs all parameters ($\beta, \theta_t, \psi$) by optimizing the shape predictor $\theta^\text{MLP}$ and the implicit shape $\psi$ using the following loss function: 
\begin{equation} \label{eq:main_objective}
\sum_{t=1}^T (
  \mathcal{L}^\text{cse}_t
  + \mathcal{L}^\text{kp}_t
  + \mathcal{L}^\text{photo}_t
  + \mathcal{L}^\text{mask}_t
  + \mathcal{L}^\text{reg}_t).
\end{equation}
The latter calculates frame-specific losses $\mathcal{L}_t^\cdot$ and sums them over all $T$ training images.
The next paragraphs detail each loss term.

\paragraph{(i)~CSE-keypoint loss.}
Besides leveraging CSE for the pose initialization, we also guide the fitting process with the following CSE keypoint loss:
\begin{equation} \label{eq:cse_keypoint_loss}
\mathcal{L}^\text{cse}_t = \sum_{\mathbf{u}_t \in M_t} \| \mathbf{u}_t - P_t F(\mathcal{V}, \beta, \theta_t)_{\text{NN}^e(\mathbf{u}_t)} \|,
\end{equation}
integrating the reprojection distances between each foreground pixel $\mathbf{u}_t \in M_t$ and the corresponding 2D projections $P_t F(\mathcal{V}, \beta, \theta_t)_{\text{NN}^e(\mathbf{u}_t)}$ of $\mathbf{u}_t$'s CSE correspondence $\text{NN}^e(\mathbf{u}_t)$ on the mesh $F(\mathcal{V}, \theta_t, \beta)_{\text{NN}^e(\mathbf{u}_t)}$.
The mesh is deformed with parameters $\theta_t = \theta^\text{MLP}(\tau(t))$ given by the pose predictor $\theta^\text{MLP}$ for time-stamp $\tau(t)$ of the frame $t$.
Here, $M_t \in \{0, 1\}^{H \times W}$ corresponds to the foreground segmentation extracted with a pre-trained segmenter \cite{carion2020end}.

\paragraph{(ii)~Sparse-keypoint loss.}
We complement the main CSE-keypoint training signal with the standard sparse-keypoint loss
$
\mathcal{L}^\text{kp}_t = \sum_{k=1}^{N_J} \| h(I_t)_k - P_t J_k \|,
$
minimizing the distance between the projection $P_t J_k$ and the detection $h(I_t)_k$ of the $j$-th SMAL joint $J_k = F(\mathcal{\hat{J}}, \beta, \theta_t)_k$ in the articulation of the image $I_t$.
Here $h(I_t) \in \mathbb{R}^{N_J \times 2}$ denotes the 2D keypoint detections from BARC's sparse dog keypoint detector \cite{rueegg_barc_2022}.
We discovered that sparse keypoints improve fitting on thin body structures, such as paws, where the CSE detections are less accurate.

\paragraph{(iii)~Photometric loss.}
To train the appearance predictor, and to provide an additional supervision for fine 3D fitting, we minimize the photometric loss:
\begin{equation} \label{eq:photometric_loss}
\mathcal{L}^\text{photo}_t = 
\text{LPIPS}(
    M_t \odot I_t
    ,
    \mathcal{R}(
        P_t,
        F(\mathcal{V}, \theta_t, \beta),
        \mathcal{F},
        \psi^\text{MLP}
    )
),
\end{equation}
comprising the Learned Perceptual Image Patch Similarity~\cite{zhang2018perceptual} (LPIPS) between the masked ground-truth image $M_t \odot I_t$ and the RGB render of the posed mesh $F(\mathcal{V}, \theta_t, \beta)$ colored by the implicit duplex-mesh MLP $\psi^\text{MLP}$.

\paragraph{(iv)~Chamfer mask loss.}
Similar to~\cite{rueegg_barc_2022,ruegg_bite_2023,biggs_creatures_2018}, we minimize a silhouette loss $\mathcal{L}^\text{mask}_t$:
\begin{equation} \label{eq:mask_loss}
\mathcal{L}^\text{mask}_t = 
d_\text{Chamfer}(
    \{ \bu_t | \bu_t \in M_t \},
    \{ P_t X_k | X_k \in F(\mathcal{V}, \theta_t, \beta)\}
),
\end{equation}
evaluating Chamfer distance between the set of occupied 2D pixels $\bu_t \in M_t$ of the g. t. mask $M_t$,
and the 2D projections of the posed-mesh vertices $F(\mathcal{V}, \theta_t, \beta)$.

\paragraph{(v) Shape regularizers.}
To avoid implausible mesh articulations, we also employ shape regularization loss
$
\mathcal{L}^\text{reg}_t = 
    \mathcal{L}^\text{arap}_t
    + \mathcal{L}^\text{edge}_t
$,
comprising the As-Rigid-As-Possible (ARAP) regularizer $\mathcal{L}^\text{arap}_t$ \cite{sorkine_laplacian_2004},
and the edge-length penalty $\mathcal{L}^\text{edge}_t$ \cite{kar2015category} enforcing local rigidity of the posed template. See supplementary for details.

\begin{figure*}
    \centering
    \newcommand{\qualfigsize}{1.5cm}
\graphicspath{{figures/img_folder/qualitative}}

\newoutputstream{imagefile}
\openoutputfile{\jobname.img}{imagefile}
\newcommand{\logimage}[1]{%
  \addtostream{imagefile}{#1}%
}

\newcommand*{\onefig}[2]{%
\logimage{#1.png}%
\includegraphics[height=\qualfigsize,width=\qualfigsize,keepaspectratio,align=c,trim=0 0 0 0, clip]{#1}%
}

\newcommand*{\groupfig}[3]{%
\ifthenelse{\equal{#1}{GT}}{%
\onefig{NVS_seq_#2_gt_frame_#3}{#2}%
&%
&%
}{%
\ifthenelse{\equal{#1}{BARC} \OR \equal{#1}{BITE}}{%

\parbox{1cm}{
    \begin{tikzpicture}
        \draw (0,0) -- (1,1); %
    \end{tikzpicture}
}&

\onefig{#1_seq_#2_texture_frame_#3}{#2}%
\onefig{#1_seq_#2_texture_frame_\the\numexpr#3+30\relax}{#2}&%
}{%
\onefig{#1_seq_#2_render_frame_#3}{#2}&%
\onefig{#1_seq_#2_texture_frame_#3}{#2}%
\onefig{#1_seq_#2_render_frame_\the\numexpr#3+30\relax}{#2}&%
}}}

\newcommand*{\biggroupfigF}[5]{%
\groupfig{#1}{#2}{#3}&%
\groupfig{#1}{#2}{#4}&%
\groupfig{#1}{#2}{#5}%
}

\newcommand*{\rowlabel}[1]{%
\scriptsize%
#1%
}

\renewcommand{\arraystretch}{0}
\setlength{\tabcolsep}{0pt}
\centering%

\footnotesize%
\resizebox{\linewidth}{!}{%
\begin{tabular}{cccccccccccccccc}%
\rowlabel{{G.T.}}&\biggroupfigF{GT}{1009_3157_2796}{0}{50}{150}\\%
\rowlabel{{BARC}}&\biggroupfigF{BARC}{1009_3157_2796}{0}{50}{150}\\%
\rowlabel{{BITE}}&\biggroupfigF{BITE}{1009_3157_2796}{0}{50}{150}\\%
\rowlabel{{RAC}}&\biggroupfigF{RAC}{1009_3157_2796}{0}{50}{150}\\
\rowlabel{{\textbf{Ours}}}&\biggroupfigF{NVS}{1009_3157_2796}{0}{50}{150}\\%
\rowlabel{{G.T.}}&\biggroupfigF{GT}{1050_43243_38827}{0}{50}{150}\\%
\rowlabel{{BARC}}&\biggroupfigF{BARC}{1050_43243_38827}{0}{50}{150}\\%
\rowlabel{{BITE}}&\biggroupfigF{BITE}{1050_43243_38827}{0}{50}{150}\\%
\rowlabel{{RAC}}&\biggroupfigF{RAC}{1050_43243_38827}{0}{50}{150}\\%
\rowlabel{{\textbf{Ours}}}&\biggroupfigF{NVS}{1050_43243_38827}{0}{50}{150}\\%
\rowlabel{{G.T.}}&\biggroupfigF{GT}{1022_15133_10069}{0}{50}{120}\\%
\rowlabel{{BARC}}&\biggroupfigF{BARC}{1022_15133_10069}{0}{50}{120}\\%
\rowlabel{{BITE}}&\biggroupfigF{BITE}{1022_15133_10069}{0}{50}{120}\\%
\rowlabel{{RAC}}&\biggroupfigF{RAC}{1022_15133_10069}{0}{50}{120}\\%
\rowlabel{{\textbf{Ours}}}&\biggroupfigF{NVS}{1022_15133_10069}{0}{50}{120}\\%
\end{tabular}%
}

    \caption{
    \textbf{Qualitative comparison}. 
    We note that, unlike template-based approaches, the reconstructed meshes from RAC are very far from the actual shape of a dog. To evaluate temporal consistency, please refer to the result \webpagelink{videos}.
    }
    \label{fig:qualitative}
\end{figure*}
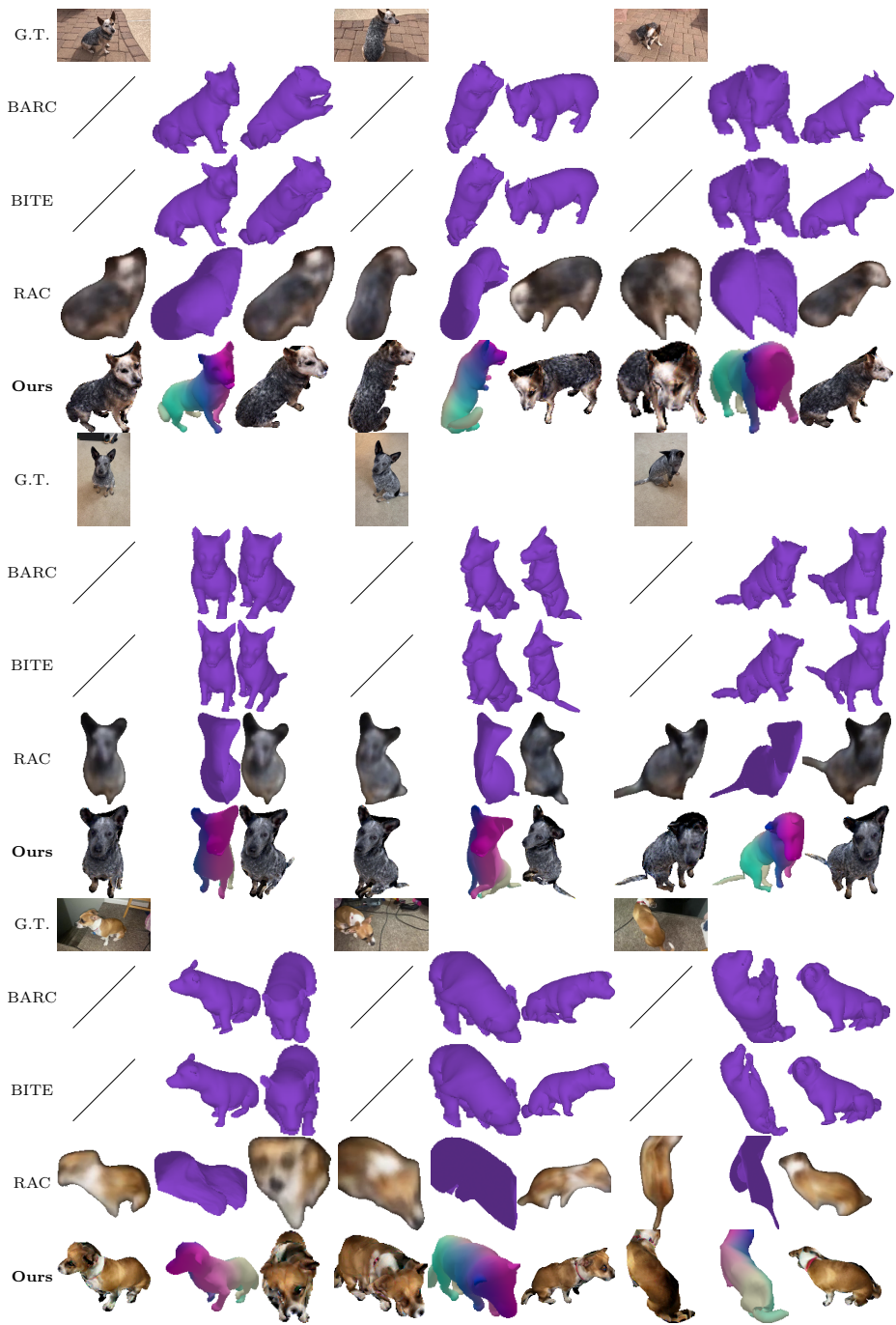

\section{Experiments} \label{sec:experiments}

\vspace*{-0.1in}
\begin{table}[b!]%
\centering%
\caption{Average results on 50 sequences from COP3D and 39 sequences from APTv2. Quality of pose estimates is measured by IoU, IoUw5, $err_\text{track}$; appearance quality is measured by PSNR, PSNRw5, LPIPS. 
Note that BARC and BITE only estimate pose and hence we cannot evaluate appearance quality, indicated by `$\times$'.}
\label{tab:main_quantitative}
\begin{tabular}{r|ccccc|c} \toprule
Dataset & \multicolumn{5}{c|}{\texttt{CoP3D}~\cite{sinha_common_2022}} & \texttt{APTv2}~\cite{yang2023aptv2} \\ \midrule
 &  IoU $\uparrow$ &  IoUw5 $\uparrow$ & PSNR $\uparrow$ & PSNRw5 $\uparrow$ & LPIPS $\downarrow$ & $err_\text{track}$ $\downarrow$ \\ \midrule
BARC \cite{rueegg_barc_2022}        & 0.75  & 0.47 & $\times$  & $\times$     & $\times$  & 0.047  \\
BITE \cite{ruegg_bite_2023}         & 0.81  & 0.59 &  $\times$  & $\times$     & $\times$ & 0.047   \\
RAC \cite{yang_reconstructing_2023} & 0.76  & 0.52 & 21.86  & 17.51     & 0.164 & 0.093    \\
Ours                                & \textbf{0.84}  & \textbf{0.79} & \textbf{22.12}  & \textbf{19.40}     & \textbf{0.041} & \textbf{0.035} \\ \bottomrule%

\end{tabular}
\end{table}%

\paragraph{Datasets.}
We evaluate all models on COP3D~\cite{sinha_common_2022}, an open-source dataset containing %
fly-around videos of pets annotated with cameras and segmentation masks.
We select a subset of 50 dog videos with a variety of poses, movements and textures.
Each test video contains 200 frames, which we split to the training and test sets by considering contiguous blocks of 15 frames as train, interleaved by blocks of 5 frames as test (we further evaluate the impact of this protocol in \cref{tab:train_test_splits}).
We evaluate all models at 256$\times$256 resolution.

We also propose a tracking evaluation using APTv2~\cite{yang2023aptv2}, a comprehensive dataset that includes %
videos featuring 30 distinct animal species in motion. Each video frame in this dataset is annotated with 17 keypoints, which we use as ground-truth for quantitative evaluation. All videos in the dataset have a consistent length of 15 frames. We restrict to a subset of 39 sequences, where the video contains a single-instance of a dog.

\paragraph{Metrics.}
On COP3D~\cite{sinha_common_2022}, we report three evaluation metrics assessing the quality of the predicted shape and texture:
(i)~\textbf{IoU} reports the average over the intersection-over-union between the silhouette render of the posed shape $S_t$ and the ground truth mask $M_t$ computed for each frame $t$; 
(ii)~\textbf{PSNR} computes the average peak-signal-to-noise ratio between the renders $\bar{I}_t$ of the fitted shape at time $t$, and the ground truth image $I_t$; and 
(iii)~\textbf{LPIPS}~\cite{zhang2018perceptual}  is a perceptual metric that measures the average visual similarity between the rendered images $\bar{I}_t$ of the fitted shape at time $t$, and the ground truth images $I_t$. This metric is particularly useful as it takes into account human visual perception and the structural information of the images, providing a more accurate measure of visual similarity compared to PSNR. Additionally, we report \emph{worst 5\%} variant \textbf{PSNRw5}, \textbf{IoUw5}.
On APTv2~\cite{yang2023aptv2}, we evaluate $\mathbf{err_\text{track}}$ measuring the tracking error of the annotated keypoints via the following protocol: 
(i) On the first frame, we pair each ground-truth keypoint $kp_{i}^{0}$ with a vertex $v_{i} \in \mathbb{R}^{3}$ on the mesh via the predicted pose.    
(ii) On the remaining frames, we compute \emph{L2} distance between ground-truth keypoints and the projection of paired vertices.

\paragraph{Baselines.}
First, we compare to template-based reconstructors BARC \cite{rueegg_barc_2022} and BITE \cite{ruegg_bite_2023} which, similar to us, predict parameters of the SMAL model, but take as input only a single image (frame).
Here, BITE extends BARC with a test-time iterative refinement.
Secondly we compare to RAC \cite{yang_reconstructing_2023}, which is a template-free reconstructor with deformable shape and texture priors learned by observing various animals in videos.
For fair comparison, we execute their shape/texture training pipeline on the original training set extended with our CoP3D videos.

The quality of our textures can only be directly compared to RAC's model, which also includes texture, but we cannot compare them to BITE or BARC because they only output 3D shapes.

\paragraph{Results.}
\Cref{tab:main_quantitative} contains the results of our experiments on COP3D and APTv2. On COP3D, our method outperforms BARC,BITE and RAC in the quality of the predicted texture (LPIPS). In terms of IoU, we outperform BARC and RAC, and achieve similar performance to BITE. However, since our method is temporally-consistent, we significantly outperform BITE on worst 5\% IoU.
On APTv2~\cite{yang2023aptv2}, our method outperforms BARC, BITE and RAC on the tracking evaluation. We note that RAC achieves significantly worst than all the other methods. We argue that this is caused by the poor quality of the rendered shape.
We provide qualitative comparison in \cref{fig:qualitative} and \webpagelink{videos} for visual evaluation.

\begin{figure}[t]
    \centering
\includegraphics[width=.9\columnwidth]{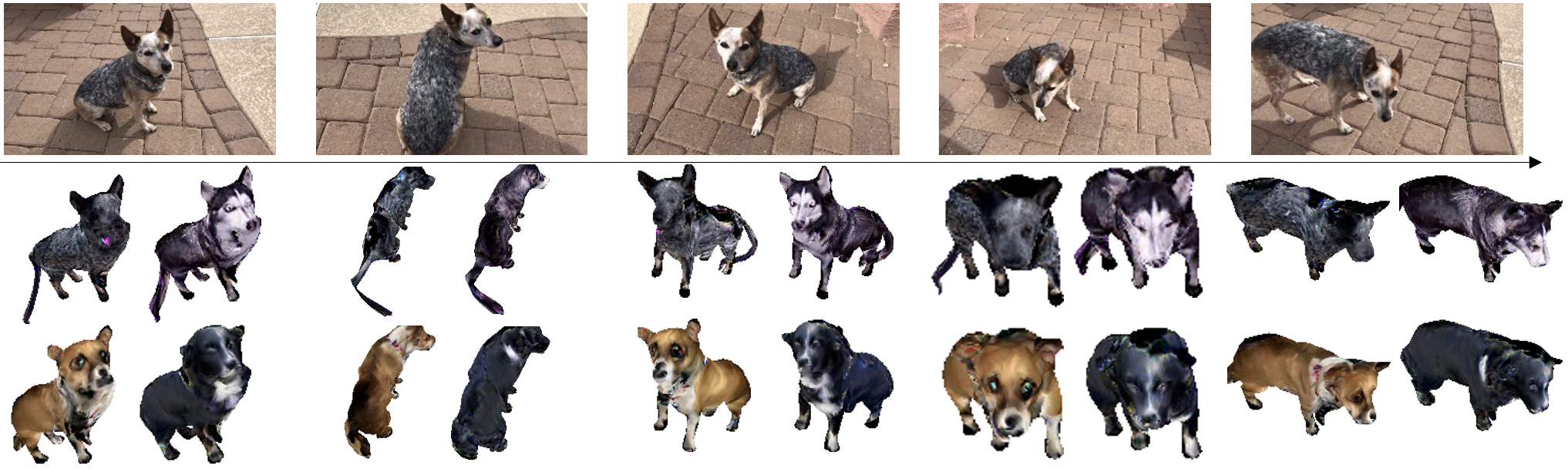}
    \caption{%
    \textbf{Texture transfer.}
    Models optimized on a scene can be re-animated via the template articulations. We demonstrate shape and motion projection from one optimized scene to other textures.}
    \label{fig:motion reprojection}
\end{figure}

\begin{table}[b!]
\small%
\centering%
\caption{
\textbf{Ablation on CoP3D} reporting performance with various loss terms removed and without camera motion factorization ($g_t^\text{cam}=g^0_t$).}
\begin{tabular}{rrrrrrrr|r}\toprule
w/o & $\mathcal{L}^\text{chamfer}$ & $\mathcal{L}^\text{cse}$ & $\mathcal{L}^\text{keypoint}$ & $\mathcal{L}^\text{color}$ & $\mathcal{L}^\text{arap}$ & $\mathcal{L}^\text{edge}$ & 
$g_t^\text{cam}=g^0_t$
& Ours \\ \midrule
IoU $\uparrow$ & 0.70 & 0.81 & 0.80 & 0.81 & 0.81 & 0.83 & 0.72 & \textbf{0.84} \\ 
PSNR $\uparrow$ & 20.65 & 20.89 & 21.54 & 21.62 & 21.61 & 21.88 & 19.12 & \textbf{22.40} \\ 
LPIPS $\downarrow$ & 0.060 & 0.051 & 0.048 & 0.047 & 0.047 & 0.045 & 0.067 &\textbf{0.038} \\ \bottomrule
\end{tabular}
\label{tab:abl_loss}
\end{table}

\paragraph{Ablation study.} 
To validate our design choices, we ablate individual components of our model and record the incurred changes in performance. 
\textit{{(i) Loss ablation.}}
In \cref{tab:abl_loss}, we remove each loss term of Equation~\ref{eq:main_objective} and report the resulting PSNR/LPIPS/IoU.
The results indicate a performance drop when any loss is removed, which confirms the merit of each loss.
\textit{{(ii) Motion factorization.}}
We also demonstrate the benefits of our factorization of measured rigid motion to the motion of the camera and the motion of the shape (\cref{sec:shape_fitting}).
Specifically, we design an experiment where the rigid motion $g_t$ of each rendering camera $P_t$ is represented with the root rigid component $g^0_t$ of the SMAL deformation coefficients $\theta_t$, i.e., $\forall t \in [1..T] g_t^\text{cam}=g^0_t$, ignoring the SfM estimate $g_t^\text{SfM}$.
Results in \cref{tab:abl_loss} indicate that this leads to a significant decrease in performance across all metrics justifying our rigid motion factorization.

\paragraph{Varying frame-split difficulty}
\label{par:cop3d_evaluation_split} For COP3D evaluations~\cite{sinha_common_2022}, we follow the original train/test split protocol (contiguous blocks of 15 frames as train, interleaved by blocks of 5 frames as test).
Removing excessive visible frames would render the problem unsolvable (e.g. it is impossible to guess the exact motion of a dog's legs given two boundary frames that are too far apart). Regardless, in \cref{tab:train_test_splits} we evaluate the impact of reducing the supervision with two additional splits.
Our method outperforms RAC on all splits, thus showing stronger robustness to weaker supervision.

\begin{table}[h!]%
\scriptsize%
\centering%
\caption{Evaluation with a varying train/test frame split on COP3D. Note that our method consistently beats RAC across a range of train/test splits. }%
\begin{tabular}{c | r | ccccc}\toprule
 Split Train/Test &&  IoU $\uparrow$ & IoUw5 $\uparrow$ & PSNR $\uparrow$ & PSNRw5 $\uparrow$ & LPIPS $\downarrow$  \\ \midrule
15/5 & RAC \cite{yang_reconstructing_2023}  & 0.76  & 0.52 & 21.86  & 17.51     & 0.164  \\
    & Ours                                & \textbf{0.84}  & \textbf{0.79} & \textbf{22.12}  & \textbf{19.40}     & \textbf{0.041} \\ \midrule
15/10 & RAC \cite{yang_reconstructing_2023} & 0.71 & 0.4  & 21.54 & 16.46 & 0.175 \\
     & Ours                                 &   \textbf{0.82} & \textbf{0.69} & \textbf{22.07} & \textbf{18.54} & \textbf{0.048} \\ \midrule
15/15 & RAC \cite{yang_reconstructing_2023} & 0.66 & 0.37 & 20.67 & 15.38 & 0.196 \\
     & Ours                                 &   \textbf{0.81} & \textbf{0.63} & \textbf{21.62} & \textbf{17.81} & \textbf{0.056} \\ \bottomrule%
\end{tabular}\label{tab:train_test_splits}%
\vspace{-0.1in}
\end{table}%

\section{Conclusions}
In this paper we proposed a novel method for generating textured animatable 3D mesh models given a casually captured monocular video of a dog.
Our method augments the animatable SMAL mesh template with Continuous Surface Embeddings to setup a surface coordinate system which, in combination with a pretrained image-to-CSE predictor, allows to estimate accurate image-to-mesh correspondences that eventually lead to significantly more accurate fits.
The better shape fitting then enables us to optimize an implicit opacity-color texture supported by a scaffold defined by the mesh in its rest pose. Our experiments reveal performance superior to existing template-free and template-based approaches on the challenging CoP3D dataset.

\newpage
\title{-- Supplementary material --}
\author{}
\institute{}
\maketitle
\label{sec:supplementary}

\section{Additional results}

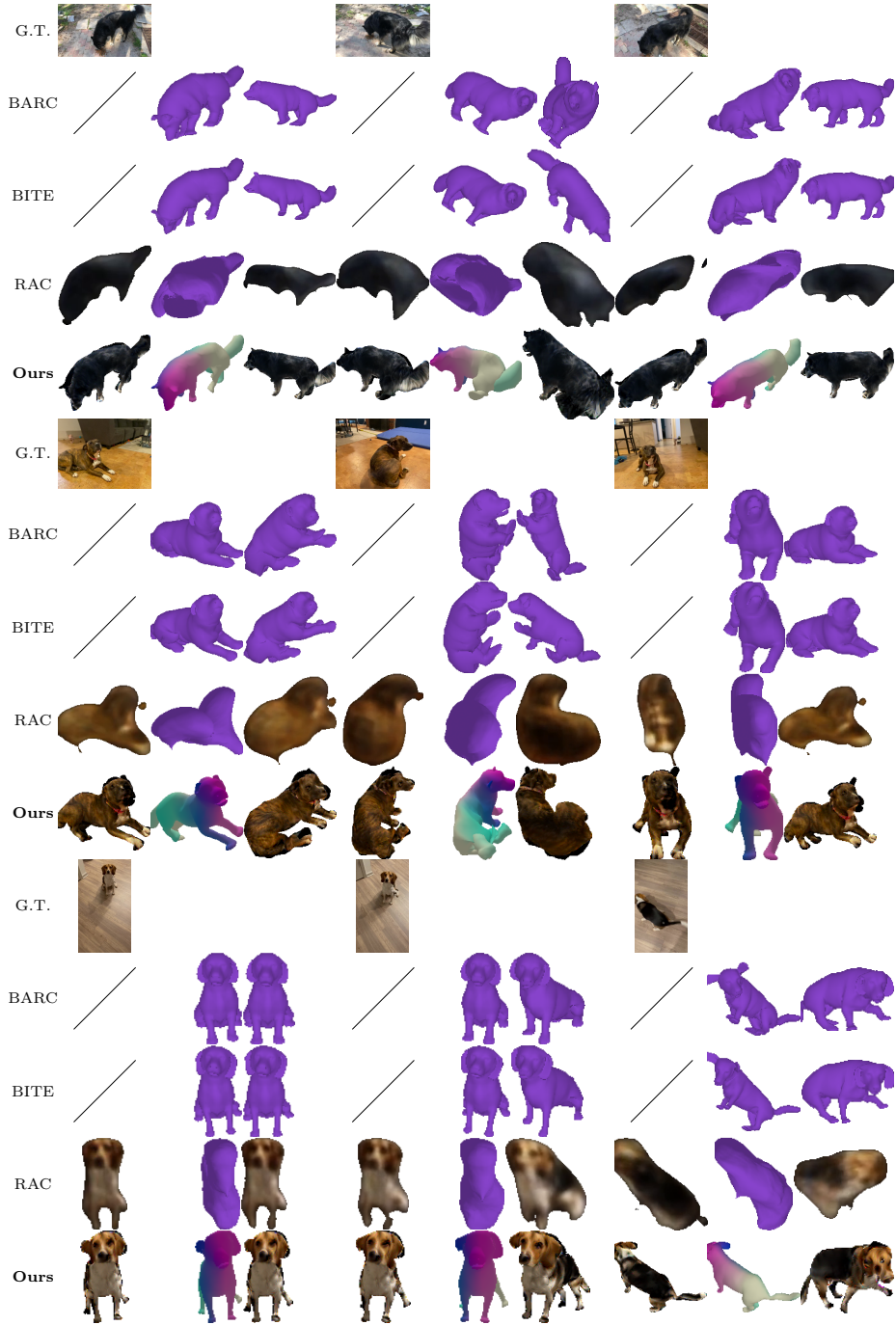
\begin{figure*}
    \centering
    \newcommand{\qualfigsize}{1.5cm}
\graphicspath{{figures/img_folder/qualitative_supplemental}}

\newcommand{\logimage}[1]{%
  \addtostream{imagefile}{#1}%
}

\newcommand*{\onefig}[2]{%
\logimage{#1.png}%
\includegraphics[height=\qualfigsize,width=\qualfigsize,keepaspectratio,align=c,trim=0 0 0 0, clip]{#1}%
}

\newcommand*{\groupfig}[3]{%
\ifthenelse{\equal{#1}{GT}}{%
\onefig{NVS_seq_#2_gt_frame_#3}{#2}%
&%
&%
}{%
\ifthenelse{\equal{#1}{BARC} \OR \equal{#1}{BITE}}{%

\parbox{1cm}{
    \begin{tikzpicture}
        \draw (0,0) -- (1,1); %
    \end{tikzpicture}
}&

\onefig{#1_seq_#2_texture_frame_#3}{#2}%
\onefig{#1_seq_#2_texture_frame_\the\numexpr#3+30\relax}{#2}&%
}{%
\onefig{#1_seq_#2_render_frame_#3}{#2}&%
\onefig{#1_seq_#2_texture_frame_#3}{#2}%
\onefig{#1_seq_#2_render_frame_\the\numexpr#3+30\relax}{#2}&%
}}}

\newcommand*{\biggroupfigF}[5]{%
\groupfig{#1}{#2}{#3}&%
\groupfig{#1}{#2}{#4}&%
\groupfig{#1}{#2}{#5}%
}

\newcommand*{\rowlabel}[1]{%
\scriptsize%
#1%
}

\renewcommand{\arraystretch}{0}
\setlength{\tabcolsep}{0pt}
\centering%

\footnotesize%
\resizebox{\linewidth}{!}{%
\begin{tabular}{cccccccccccccccc}%
\rowlabel{{G.T.}}&\biggroupfigF{GT}{1046_39156_34275}{0}{50}{150}\\%
\rowlabel{{BARC}}&\biggroupfigF{BARC}{1046_39156_34275}{0}{50}{150}\\%
\rowlabel{{BITE}}&\biggroupfigF{BITE}{1046_39156_34275}{0}{50}{150}\\%
\rowlabel{{RAC}}&\biggroupfigF{RAC}{1046_39156_34275}{0}{50}{150}\\
\rowlabel{{\textbf{Ours}}}&\biggroupfigF{NVS}{1046_39156_34275}{0}{50}{150}\\%
\rowlabel{{G.T.}}&\biggroupfigF{GT}{1037_30410_23055}{0}{50}{150}\\%
\rowlabel{{BARC}}&\biggroupfigF{BARC}{1037_30410_23055}{0}{50}{150}\\%
\rowlabel{{BITE}}&\biggroupfigF{BITE}{1037_30410_23055}{0}{50}{150}\\%
\rowlabel{{RAC}}&\biggroupfigF{RAC}{1037_30410_23055}{0}{50}{150}\\%
\rowlabel{{\textbf{Ours}}}&\biggroupfigF{NVS}{1037_30410_23055}{0}{50}{150}\\%
\rowlabel{{G.T.}}&\biggroupfigF{GT}{1043_36548_32210}{0}{50}{150}\\%
\rowlabel{{BARC}}&\biggroupfigF{BARC}{1043_36548_32210}{0}{50}{150}\\%
\rowlabel{{BITE}}&\biggroupfigF{BITE}{1043_36548_32210}{0}{50}{150}\\%
\rowlabel{{RAC}}&\biggroupfigF{RAC}{1043_36548_32210}{0}{50}{150}\\%
\rowlabel{{\textbf{Ours}}}&\biggroupfigF{NVS}{1043_36548_32210}{0}{50}{150}\\%
\end{tabular}%
}

    \caption{\textbf{Additional qualitative comparison}.}
    \label{fig:qualitative_supp2}
\end{figure*}

We include additional qualitative results in Figures \ref{fig:qualitative_supp1} and \ref{fig:qualitative_supp2}. We recommend looking at the \webpagelink{video results} to assess the quality of our reconstruction compared to other methods.
Notably, we observe that:
(1) Our reconstruction is consistent in time, which is not the case of BITE and BARC which reconstruct frames independently.
(2) The meshes reconstructed with RAC achieve relatively good score across all metrics, except LPIPS, yet they are qualitatively very poor, and they fall short in accurately representing the subject's true shape.

\section{Implicit duplex-mesh rendering}

We provide additional details for our implicit duplex-mesh model. During the rendering of $\bar{I} \in [0, 1]^{3 \times H \times W}$ from an arbitrary camera viewpoint $P$, we explained earlier that for each pixel $\bu \in [1..H] \times [1..W]$:

\begin{enumerate}
    \item We compute ${\bx}_\bu^\uparrow$, the first intersection between the ray in posed-space $\br_\bu$ and the posed outer boundary mesh $F(\hat{\mathcal{V}}^\uparrow, \beta, \theta)$. We repeat the same for the inner mesh $F(\hat{\mathcal{V}}^\uparrow, \beta, \theta)$ to get intersection point ${\bx}^\downarrow_\bu$.
    \item Since ${\bx}_\bu^\uparrow$ and ${\bx}^\downarrow_\bu$ are defined on the posed outer and inner mesh, we can compute their corresponding position $\hat{\bx}_\bu^\uparrow$ and $\hat{\bx}^\downarrow_\bu$ in the canonical space.
    \item We march in the canonical space with Emission-Absorption (EA) over the canonical ray $\hat{\br}_\bu$, defined between the two canonical points: $\hat{\bx}_\bu^\uparrow$ and $\hat{\bx}^\downarrow_\bu$
\end{enumerate}

\paragraph{Trajectory sampling} Given a ray in posed-space $\br_\bu$, the computation of (${\bx}_\bu^\uparrow$, ${\bx}^\downarrow_\bu$) can lead to 3 scenarios:
\begin{itemize}
    \item $\br_\bu$ first intersects the outer and then the inner mesh. This is the standard scenario.
    \item $\br_\bu$ only intersect the outer, and not the inner mesh. In that case, we compute ${\bx}_\bu^\uparrow$ as usual. However, since ${\bx}^\downarrow_\bu$ cannot be defined on the inner mesh, we replace it with the second intersection point between $\br_\bu$ and  $F(\hat{\mathcal{V}}^\uparrow, \beta, \theta)$. 
    \item $\br_\bu$ does not intersect with the inner or the outer mesh. Here, we do not render the pixel $\bu$.
\end{itemize}

Additionally, even if the inner and outer mesh do not overlap in the canonical space, they may intersect in some cases when they are posed. This can lead to unexpected scenarios where $\br_\bu$ first intersect the inner mesh and then the outer mesh, or only the inner mesh. We simply decide to discard these points from the rendering.

\paragraph{Optimization} 
In the method section, we described how our approach simultaneously learns the shape $\beta$, motion $\theta_{t}$, and texture $\psi$ of a sequence. During the optimization process, the implicit duplex-mesh model is randomly initialized and progressively refined via the photometric loss $\mathcal{L}^\text{photo}_t$. This loss is also backpropagated to the pose predictor. We note that the quality of texture reconstruction is highly dependent on the quality of the pose reconstruction.

\section{Shape regularizers}

We include details about the different regularizers applied on the shape during optimization.

\paragraph{As-Rigid-As-Possible (ARAP)} In the context of shape manipulation and deformation in computer graphics and geometric modeling, ARAP regularization has emerged as a pivotal technique. The main objective is to preserve the local rigidity of structures while allowing for global transformations, making it ideal for applications where maintaining the original characteristics of the shape is crucial. ARAP regularization achieves this by minimizing the deviation from a rigid transformation at a local level, effectively balancing between rigidity and flexibility.

In the specific context of our project, we observe that balancing the ARAP loss is very important to achieve a harmonious blend of realism and flexibility, and ensure that the dog's movements are both natural and within the bounds of plausible deformation. When the loss is set too high, it results in an overly rigid dog model, exhibiting minimal motion and lacking fluidity. This rigidity, while preserving local structure, inadvertently leads to an unnatural appearance in motion sequences. Conversely, setting the ARAP loss too low often lead to invalid deformations, where the mesh of the dog may contort or distort in unrealistic ways. 

We follow implementation defined in \cite{sorkine_laplacian_2004}. Please refer to the original paper for the implementation details.

\paragraph{Edge regularization} We include a regularization on the edge length between connected vertices of the mesh $\mathcal{V}$ after the mesh is posed. 

The principle behind the edge length regularizer is to minimize the variation in the length of each edge during the deformation process (when the mesh is posed). For a given edge connecting vertices \( v_i \) and \( v_j \), the length of this edge can be denoted as \( l_{ij} \). The goal is to keep \( l_{ij} \) as close as possible to its original length in canonical pose \( l_{ij}^\text{canonical} \). Mathematically, this can be formulated as a loss:
\[
\mathcal{L}^\text{edge}_t = \sum_{(i, j) \in \text{Edges}(\mathcal{F})} \left( l_{ij} - l_{ij}^\text{canonical} \right)^2
\]
By tuning the weight of the edge-length regularizer, we observe the same behaviour as the ARAP loss.  

\section{Independent Duplex-Mesh renderer}
\begin{figure}[t]
    \centering
\includegraphics[width=.9\columnwidth]{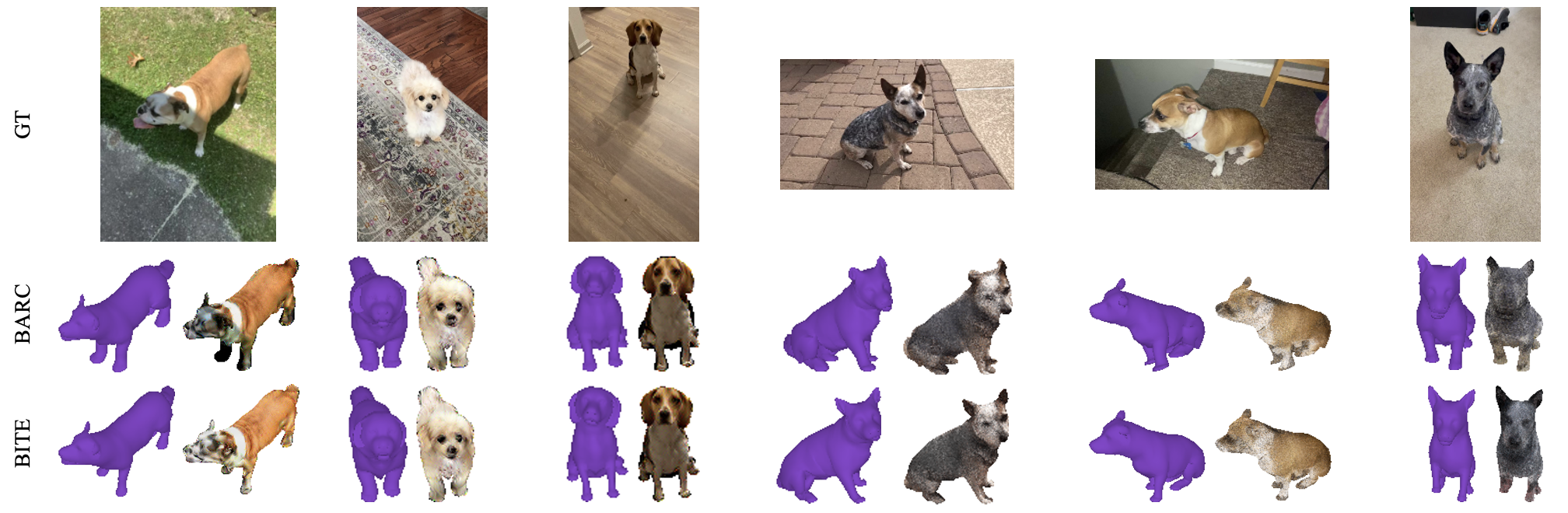}
    \caption{%
    \textbf{Texture reconstruction from BARC BITE pose predictions}. We optimize our duplex-mesh renderer model with fixed set of pose $\{\theta_{t}\}$ estimated by BARC and BITE on sequences from COP3D.}
    \label{fig:barc_bite_text_reconstruct}
\end{figure}

We emphasize that our texture rendering module is \textbf{independent} from our pose prediction framework, and can be applied on fixed pose predictions $\{\theta_{t}\}$ by optimizing a photometric loss. 
We propose to extend BARC and BITE method (which only predict pose) with our texture module, and show qualitative results in \Cref{fig:barc_bite_text_reconstruct}. Our texture model achieves strong reconstruction even on BARC and BITE predictions. In practice, since BARC and BITE generate non-temporally consistent pose, we observe a drop in texture quality for scene with substantial motions from the camera or the subject.

\begin{figure*}
    \centering
    \newcommand{\qualfigsize}{1.5cm}
\graphicspath{{figures/img_folder/qualitative_supplemental}}

\newcommand{\logimage}[1]{%
  \addtostream{imagefile}{#1}%
}

\newcommand*{\onefig}[2]{%
\logimage{#1.png}%
\includegraphics[height=\qualfigsize,width=\qualfigsize,keepaspectratio,align=c,trim=0 0 0 0, clip]{#1}%
}

\newcommand*{\groupfig}[3]{%
\ifthenelse{\equal{#1}{GT}}{%
\onefig{NVS_seq_#2_gt_frame_#3}{#2}%
&%
&%
}{%
\ifthenelse{\equal{#1}{BARC} \OR \equal{#1}{BITE}}{%

\parbox{1cm}{
    \begin{tikzpicture}
        \draw (0,0) -- (1,1); %
    \end{tikzpicture}
}&

\onefig{#1_seq_#2_texture_frame_#3}{#2}%
\onefig{#1_seq_#2_texture_frame_\the\numexpr#3+30\relax}{#2}&%
}{%
\onefig{#1_seq_#2_render_frame_#3}{#2}&%
\onefig{#1_seq_#2_texture_frame_#3}{#2}%
\onefig{#1_seq_#2_render_frame_\the\numexpr#3+30\relax}{#2}&%
}}}

\newcommand*{\biggroupfigF}[5]{%
\groupfig{#1}{#2}{#3}&%
\groupfig{#1}{#2}{#4}&%
\groupfig{#1}{#2}{#5}%
}

\newcommand*{\rowlabel}[1]{%
\scriptsize%
#1%
}

\renewcommand{\arraystretch}{0}
\setlength{\tabcolsep}{0pt}
\centering%

\footnotesize%
\resizebox{\linewidth}{!}{%
\begin{tabular}{cccccccccccccccc}%
\rowlabel{{G.T.}}&\biggroupfigF{GT}{1048_41121_36801}{0}{50}{150}\\%
\rowlabel{{BARC}}&\biggroupfigF{BARC}{1048_41121_36801}{0}{50}{150}\\%
\rowlabel{{BITE}}&\biggroupfigF{BITE}{1048_41121_36801}{0}{50}{150}\\%
\rowlabel{{RAC}}&\biggroupfigF{RAC}{1048_41121_36801}{0}{50}{150}\\
\rowlabel{{\textbf{Ours}}}&\biggroupfigF{NVS}{1048_41121_36801}{0}{50}{150}\\%
\rowlabel{{G.T.}}&\biggroupfigF{GT}{1010_4414_3789}{0}{50}{150}\\%
\rowlabel{{BARC}}&\biggroupfigF{BARC}{1010_4414_3789}{0}{50}{150}\\%
\rowlabel{{BITE}}&\biggroupfigF{BITE}{1010_4414_3789}{0}{50}{150}\\%
\rowlabel{{RAC}}&\biggroupfigF{RAC}{1010_4414_3789}{0}{50}{150}\\%
\rowlabel{{\textbf{Ours}}}&\biggroupfigF{NVS}{1010_4414_3789}{0}{50}{150}\\%
\rowlabel{{G.T.}}&\biggroupfigF{GT}{565_81664_160332}{0}{50}{150}\\%
\rowlabel{{BARC}}&\biggroupfigF{BARC}{565_81664_160332}{0}{50}{150}\\%
\rowlabel{{BITE}}&\biggroupfigF{BITE}{565_81664_160332}{0}{50}{150}\\%
\rowlabel{{RAC}}&\biggroupfigF{RAC}{565_81664_160332}{0}{50}{150}\\%
\rowlabel{{\textbf{Ours}}}&\biggroupfigF{NVS}{565_81664_160332}{0}{50}{150}\\%
\end{tabular}%
}

    \caption{\textbf{Additional qualitative comparison}.}
    \label{fig:qualitative_supp1}
\end{figure*}

\newpage
\bibliographystyle{splncs04}
\bibliography{refs/main}

\begin{thebibliography}{10}
\providecommand{\url}[1]{\texttt{#1}}
\providecommand{\urlprefix}{URL }
\providecommand{\doi}[1]{https://doi.org/#1}

\bibitem{badger2020}
Badger, M., Wang, Y., Modh, A., Perkes, A., Kolotouros, N., Pfrommer, B., Schmidt, M., Daniilidis, K.: {3D} bird reconstruction: a dataset, model, and shape recovery from a single view. In: Eur. Conf. Comput. Vis. (2020)

\bibitem{biggs_creatures_2018}
Biggs, B., Roddick, T., Fitzgibbon, A., Cipolla, R.: Creatures great and smal: {Recovering} the shape and motion of animals from video. In: Asian Conf. Comput. Vis. pp. 3--19. Springer (2018)

\bibitem{carion2020end}
Carion, N., Massa, F., Synnaeve, G., Usunier, N., Kirillov, A., Zagoruyko, S.: End-to-end object detection with transformers. In: Computer Vision – ECCV 2020: 16th European Conference, Glasgow, UK, August 23–28, 2020, Proceedings, Part I. p. 213–229. Springer-Verlag (2020)

\bibitem{chan2022efficient}
Chan, E.R., Lin, C.Z., Chan, M.A., Nagano, K., Pan, B., De~Mello, S., Gallo, O., Guibas, L.J., Tremblay, J., Khamis, S., et~al.: Efficient geometry-aware 3d generative adversarial networks. In: Proceedings of the {IEEE}/{CVF} Conference on Computer Vision and Pattern Recognition. pp. 16123--16133 (2022)

\bibitem{chen2021animatable}
Chen, J., Zhang, Y., Kang, D., Zhe, X., Bao, L., Jia, X., Lu, H.: Animatable neural radiance fields from monocular rgb videos (2021)

\bibitem{dong2023ag3d}
Dong, Z., Chen, X., Yang, J., Black, M.J., Hilliges, O., Geiger, A.: {AG3D}: {L}earning to {G}enerate {3D} {A}vatars from {2D} {I}mage {C}ollections. In: Int. Conf. Comput. Vis. (2023)

\bibitem{ehsani_who_2018}
Ehsani, K., Bagherinezhad, H., Redmon, J., Mottaghi, R., Farhadi, A.: Who let the dogs out? modeling dog behavior from visual data. IEEE Conf. Comput. Vis. Pattern Recog. pp. 4051--4060 (2018)

\bibitem{gao2021dynamic}
Gao, C., Saraf, A., Kopf, J., Huang, J.B.: Dynamic view synthesis from dynamic monocular video. In: Int. Conf. Comput. Vis. (2021)

\bibitem{goel_humans_2023}
Goel, S., Pavlakos, G., Rajasegaran, J., Kanazawa, A., Malik, J.: Humans in 4{D}: Reconstructing and tracking humans with transformers. In: Int. Conf. Comput. Vis. (2023)

\bibitem{gu_ava_2018}
Gu, C., Sun, C., Vijayanarasimhan, S., Pantofaru, C., Ross, D.A., Toderici, G., Li, Y., Ricco, S., Sukthankar, R., Schmid, C., Malik, J.: Ava: A video dataset of spatio-temporally localized atomic visual actions. IEEE Conf. Comput. Vis. Pattern Recog. pp. 6047--6056 (2017)

\bibitem{guo2023vid2avatar}
Guo, C., Jiang, T., Chen, X., Song, J., Hilliges, O.: Vid2avatar: 3d avatar reconstruction from videos in the wild via self-supervised scene decomposition. In: IEEE Conf. Comput. Vis. Pattern Recog. (June 2023)

\bibitem{ionescu_human36m_2014}
Ionescu, C., Papava, D., Olaru, V., Sminchisescu, C.: Human3.6m: Large scale datasets and predictive methods for 3d human sensing in natural environments. IEEE Trans. Pattern Anal. Mach. Intell.  \textbf{36},  1325--1339 (2014)

\bibitem{jiang2022neuman}
Jiang, W., Yi, K.M., Samei, G., Tuzel, O., Ranjan, A.: Neuman: Neural human radiance field from a single video. In: Eur. Conf. Comput. Vis. (2022)

\bibitem{kar2015category}
Kar, A., Tulsiani, S., Carreira, J., Malik, J.: Category-specific object reconstruction from a single image. IEEE Conf. Comput. Vis. Pattern Recog. pp. 1966--1974 (2014)

\bibitem{kearney_rgbd-dog_2020}
Kearney, S., Li, W., Parsons, M., Kim, K.I., Cosker, D.: Rgbd-dog: Predicting canine pose from rgbd sensors. In: IEEE Conf. Comput. Vis. Pattern Recog. (June 2020)

\bibitem{khosla_novel_2011}
Khosla, A., Jayadevaprakash, N., Yao, B., Fei-Fei, L.: Novel dataset for fine-grained image categorization. In: First Workshop on Fine-Grained Visual Categorization, IEEE Conference on Computer Vision and Pattern Recognition. Colorado Springs, CO (June 2011)

\bibitem{Kokkinos_2021_CVPR}
Kokkinos, F., Kokkinos, I.: Learning monocular 3d reconstruction of articulated categories from motion. In: IEEE Conf. Comput. Vis. Pattern Recog. pp. 1737--1746 (June 2021)

\bibitem{efficientPnp_09}
Lepetit, V., Moreno-Noguer, F., Fua, P.: Epnp: An accurate o(n) solution to the pnp problem. International Journal Of Computer Vision  \textbf{81},  155--166 (2009)

\bibitem{lewis2023pose}
Lewis, J.P., Cordner, M., Fong, N.: Pose Space Deformation: A Unified Approach to Shape Interpolation and Skeleton-Driven Deformation. Association for Computing Machinery, New York, NY, USA, 1 edn. (2023)

\bibitem{li2021coarsetofine}
Li, C., Lee, G.H.: Coarse-to-fine animal pose and shape estimation. ArXiv  \textbf{abs/2111.08176} (2021)

\bibitem{lin_microsoft_2015}
Lin, T.Y., Maire, M., Belongie, S.J., Hays, J., Perona, P., Ramanan, D., Dollár, P., Zitnick, C.L.: Microsoft {COCO}: {Common} {Objects} in {Context}. In: Eur. Conf. Comput. Vis. (2014)

\bibitem{liu_soft_2019}
Liu, S., Li, T., Chen, W., Li, H.: Soft rasterizer: A differentiable renderer for image-based 3d reasoning. Int. Conf. Comput. Vis.  (Oct 2019)

\bibitem{lombardi_neural_2019}
Lombardi, S., Simon, T., Saragih, J., Schwartz, G., Lehrmann, A., Sheikh, Y.: Neural volumes: Learning dynamic renderable volumes from images. ACM Trans. Graph.  \textbf{38}(4),  65:1--65:14 (Jul 2019)

\bibitem{loper_smpl_2015}
Loper, M., Mahmood, N., Romero, J., Pons-Moll, G., and, M.J.B.: {SMPL}: {A} skinned multi- person linear model. ACM Trans. Graph.  (2015)

\bibitem{mehta_monocular_2017}
Mehta, D., Rhodin, H., Casas, D., Fua, P., Sotnychenko, O., Xu, W., Theobalt, C.: Monocular {3D} {Human} {Pose} {Estimation} in the {Wild} {Using} {Improved} {CNN} {Supervision}. In: {3DV} (2017)

\bibitem{melzi_zoomout_2019}
Melzi, S., Ren, J., Rodolà, E., Sharma, A., Wonka, P., Ovsjanikov, M.: {ZoomOut}: spectral upsampling for efficient shape correspondence. ACM Trans. Graph.  \textbf{38}(6),  155:1--155:14 (2019)

\bibitem{mildenhall_nerf_2020}
Mildenhall, B., Srinivasan, P.P., Tancik, M., Barron, J.T., Ramamoorthi, R., Ng, R.: Nerf: {Representing} scenes as neural radiance fields for view synthesis. Eur. Conf. Comput. Vis.  (2020)

\bibitem{neverova_continuous_2020}
Neverova, N., Novotný, D., Vedaldi, A.: Continuous {Surface} {Embeddings}. In: Adv. Neural Inform. Process. Syst. (2020)

\bibitem{novotny2022keytr}
Novotny, D., Rocco, I., Sinha, S., Carlier, A., Kerchenbaum, G., Shapovalov, R., Smetanin, N., Neverova, N., Graham, B., Vedaldi, A.: Keytr: keypoint transporter for 3d reconstruction of deformable objects in videos. In: IEEE Conf. Comput. Vis. Pattern Recog. pp. 5595--5604 (2022)

\bibitem{osman_star_2020}
Osman, A.A.A., Bolkart, T., Black, M.J.: {STAR}: A sparse trained articulated human body regressor. In: Eur. Conf. Comput. Vis. pp. 598--613 (2020)

\bibitem{pavlakos_coarse--fine_2017}
Pavlakos, G., Zhou, X., Derpanis, K.G., Daniilidis, K.: Coarse-to-fine volumetric prediction for single-image 3{D} human pose. In: IEEE Conf. Comput. Vis. Pattern Recog. (2017)

\bibitem{peng2021animatable}
Peng, S., Dong, J., Wang, Q., Zhang, S., Shuai, Q., Zhou, X., Bao, H.: Animatable neural radiance fields for modeling dynamic human bodies. In: Int. Conf. Comput. Vis. pp. 14294--14303 (2021)

\bibitem{pumarola2021d}
Pumarola, A., Corona, E., Pons-Moll, G., Moreno-Noguer, F.: {D-NeRF: Neural Radiance Fields for Dynamic Scenes}. In: IEEE Conf. Comput. Vis. Pattern Recog. (2020)

\bibitem{ravi_accelerating_2020}
Ravi, N., Reizenstein, J., Novotny, D., Gordon, T., Lo, W.Y., Johnson, J., Gkioxari, G.: Accelerating {3D} {Deep} {Learning} with {PyTorch3D}. arXiv  (2020)

\bibitem{romero_embodied_2017}
Romero, J., Tzionas, D., Black, M.J.: Embodied hands: modeling and capturing hands and bodies together. ACM Trans. Graph.  (2017)

\bibitem{rueegg_barc_2022}
Rueegg, N., Zuffi, S., Schindler, K., Black, M.J.: {BARC}: Learning to regress {3D} dog shape from images by exploiting breed information. In: IEEE Conf. Comput. Vis. Pattern Recog. pp. 3876--3884 (2022)

\bibitem{ruegg_bite_2023}
R\"uegg, N., Tripathi, S., Schindler, K., Black, M.J., Zuffi, S.: {BITE}: Beyond priors for improved three-{D} dog pose estimation. In: IEEE Conf. Comput. Vis. Pattern Recog. pp. 8867--8876 (Jun 2023)

\bibitem{schonberger2016structure}
Schönberger, J.L., Frahm, J.M.: Structure-from-{Motion} {Revisited}. In: IEEE Conf. Comput. Vis. Pattern Recog. (2016)

\bibitem{sinha_common_2022}
Sinha, S., Shapovalov, R., Reizenstein, J., Rocco, I., Neverova, N., Vedaldi, A., Novotny, D.: Common pets in 3d: Dynamic new-view synthesis of real-life deformable categories. IEEE Conf. Comput. Vis. Pattern Recog.  (2023)

\bibitem{sorkine_laplacian_2004}
Sorkine, O., Cohen-Or, D., Lipman, Y., Alexa, M., Rössl, C., Seidel, H.P.: Laplacian {Surface} {Editing}. In: Proc. {Eurographics}. vol.~71, pp. 175--184 (2004)

\bibitem{vaswani_attention_2017}
Vaswani, A., Shazeer, N., Parmar, N., Uszkoreit, J., Jones, L., Gomez, A.N., Kaiser, L., Polosukhin, I.: Attention is all you need. Adv. Neural Inform. Process. Syst.  (2017)

\bibitem{wan2023learning}
Wan, Z., Richardt, C., Bo{\v{z}}i{\v{c}}, A., Li, C., Rengarajan, V., Nam, S., Xiang, X., Li, T., Zhu, B., Ranjan, R., et~al.: Learning neural duplex radiance fields for real-time view synthesis. In: IEEE Conf. Comput. Vis. Pattern Recog.

\bibitem{wang2021birds}
Wang, Y., Kolotouros, N., Daniilidis, K., Badger, M.: Birds of a feather: Capturing avian shape models from images. In: IEEE Conf. Comput. Vis. Pattern Recog. (2021)

\bibitem{wiles2020synsin}
Wiles, O., Gkioxari, G., Szeliski, R., Johnson, J.: Synsin: {End}-to-end view synthesis from a single image. In: IEEE Conf. Comput. Vis. Pattern Recog. pp. 7467--7477 (2020)

\bibitem{wu_ai_2019}
Wu, J., Zheng, H., Zhao, B., Li, Y., Yan, B., Liang, R., Wang, W., Zhou, S., Lin, G., Fu, Y., Wang, Y., Wang, Y.: {AI} challenger : A large-scale dataset for going deeper in image understanding. In: Int. Conf. Multimedia and Expo. pp. 1480--1485 (2019)

\bibitem{wu2023magicpony}
Wu, S., Li, R., Jakab, T., Rupprecht, C., Vedaldi, A.: {MagicPony}: Learning articulated 3d animals in the wild. In: IEEE Conf. Comput. Vis. Pattern Recog. (2023)

\bibitem{xu_vitpose_2023}
Xu, Y., Zhang, J., Zhang, Q., Tao, D.: Vitpose++: Vision transformer for generic body pose estimation. IEEE Trans. Pattern Anal. Mach. Intell.  \textbf{46}(02),  1212--1230 (feb 2024)

\bibitem{yang_lasr_2021}
Yang, G., Sun, D., Jampani, V., Vlasic, D., Cole, F., Chang, H., Ramanan, D., Freeman, W.T., Liu, C.: {LASR}: {Learning} {Articulated} {Shape} {Reconstruction} from a {Monocular} {Video}. In: IEEE Conf. Comput. Vis. Pattern Recog. (2021), \_eprint: 2105.02976

\bibitem{yang2021viser}
Yang, G., Sun, D., Jampani, V., Vlasic, D., Cole, F., Liu, C., Ramanan, D.: Viser: Video-specific surface embeddings for articulated 3d shape reconstruction. In: Adv. Neural Inform. Process. Syst. (2021)

\bibitem{yang_banmo_2021}
Yang, G., Vo, M., Neverova, N., Ramanan, D., Vedaldi, A., Joo, H.: Banmo: Building animatable 3d neural models from many casual videos. In: IEEE Conf. Comput. Vis. Pattern Recog. (2022)

\bibitem{yang_reconstructing_2023}
Yang, G., Wang, C., Reddy, N.D., Ramanan, D.: Reconstructing animatable categories from videos. In: IEEE Conf. Comput. Vis. Pattern Recog. (2023)

\bibitem{yang2023aptv2}
Yang, Y., Deng, Y., Xu, Y., Zhang, J.: Aptv2: Benchmarking animal pose estimation and tracking with a large-scale dataset and beyond (2023)

\bibitem{yu_pixelnerf_2020}
Yu, A., Ye, V., Tancik, M., Kanazawa, A.: {pixelNeRF}: {Neural} {Radiance} {Fields} from {One} or {Few} {Images}. Eur. Conf. Comput. Vis.  (2020)

\bibitem{zhang_pymaf-x_2023}
Zhang, H., Tian, Y., Zhang, Y., Li, M., An, L., Sun, Z., Liu, Y.: Pymaf-x: Towards well-aligned full-body model regression from monocular images. IEEE Trans. Pattern Anal. Mach. Intell.  (2023)

\bibitem{zhang2018perceptual}
Zhang, R., Isola, P., Efros, A.A., Shechtman, E., Wang, O.: The unreasonable effectiveness of deep features as a perceptual metric. In: IEEE Conf. Comput. Vis. Pattern Recog. pp. 586--595 (2018)

\bibitem{zuffi_3d_2017}
Zuffi, S., Kanazawa, A., Jacobs, D.W., Black, M.J.: 3d menagerie: Modeling the 3d shape and pose of animals. IEEE Conf. Comput. Vis. Pattern Recog. pp. 5524--5532 (2016)

\end{thebibliography}

\end{document}